\documentclass[journal]{IEEEtran}

\usepackage{cite}

\ifCLASSINFOpdf
\else
\fi

\usepackage{amsmath,amssymb,bm}

\usepackage{algorithmic}

\usepackage{array}

\ifCLASSOPTIONcompsoc
  \usepackage[caption=false,font=normalsize,labelfont=sf,textfont=sf]{subfig}
\else
  \usepackage[caption=false,font=footnotesize]{subfig}
\fi

\usepackage{url}
\usepackage{algorithm}
\usepackage{balance}
\usepackage{color}
\usepackage{epsfig}
\usepackage{epstopdf}
\usepackage{enumitem}
\usepackage{float}
\usepackage{graphicx}
\usepackage{setspace}
\usepackage{wasysym}

\setenumerate[1]{itemsep=0pt,partopsep=0pt,parsep=\parskip,topsep=0pt}

\newcommand{\red}{\color{red}}
\newcommand{\blue}{\color{blue}}
\newcommand{\green}{\color{green}}

\begin{document}
\singlespacing
\title{Perception-based energy functions in seam-cutting}

\author{Nan Li \thanks{N. Li is with the Center for Applied Mathematics, Tianjin University, Tianjin 300072, China. E-mail: nan@tju.edu.cn.},
        Tianli Liao,
        and Chao Wang
\thanks{T. Liao is with the Center for Combinatorics, Nankai University, Tianjin 300071, China. Email: liaotianli@mail.nankai.edu.cn.}
\thanks{C. Wang is with the Department of Software, Nankai University, Tianjin 300071, China. Email: wangchao@nankai.edu.cn.}}

\maketitle

\begin{abstract}
  Image stitching is challenging in consumer-level photography, due to alignment difficulties in
  unconstrained shooting environment. Recent studies show that seam-cutting approaches can effectively relieve artifacts generated by local misalignment. Normally, seam-cutting is described in terms of energy minimization, however, few of existing methods consider human perception in their energy functions, which sometimes causes that a seam with minimum energy is not most invisible in the overlapping region. In this paper, we propose a novel perception-based energy function in the seam-cutting framework, which considers the nonlinearity and the nonuniformity of human perception in energy minimization. Our perception-based approach adopts a sigmoid metric to characterize the perception of color discrimination, and a saliency weight to simulate that human eyes incline to pay more attention to salient objects. In addition, our seam-cutting composition can be easily implemented into other stitching pipelines. Experiments show that our method outperforms the seam-cutting method of the normal energy function, and a user study demonstrates that our composed results are more consistent with human perception.
\end{abstract}


\IEEEpeerreviewmaketitle

\section{Introduction}

\IEEEPARstart{I}{mage} stitching is a well studied topic in computer vision~\cite{szeliski2004image}, which mainly consists of alignment~\cite{Szeliski:1997,Brown:2007,gao2011constructing,zaragoza2013projective}, composition~\cite{peleg1981elimination,duplaquet1998building,davis1998mosaics,efros2001image,mills2009image} and blending~\cite{burt1983multiresolution,Perez:2003,levin2004seamless}. In consumer-level photography, it is difficult to achieve perfect alignment due to unconstrained shooting environment, so image composition becomes the most crucial step to produce artifacts-free results.

\begin{figure}[!t]
  \centering
  \includegraphics[width=0.45\textwidth]{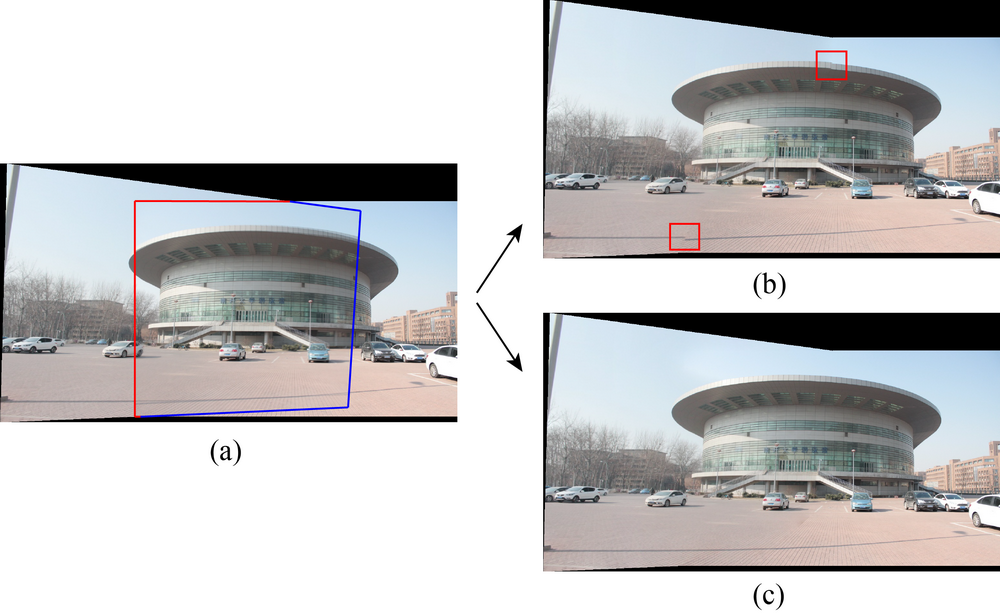}\\
  \caption{A composed result comparison between different energy functions. (a) Overlapping region.
  (b) Composed result corresponding to the normal energy function. (c) Composed result corresponding to our perception-based energy function.}
  \label{contrast_1}
\end{figure}

Seam-cutting~\cite{kwatra2003graphcut,agarwala2004interactive,Eden2006seamless,jia2008image,zhang2016multi} is a powerful composition method, which intends to find an invisible seam in the overlapping region of aligned images. Mainstream algorithms usually express the problem in terms of energy minimization and minimize it via graph-cut optimization~\cite{boykov2001fast,boykov2004experimental,kolmogorov2004energy}. Normally, for a given overlapping region of aligned images, different energy functions correspond to different seams, and certainly correspond to different composed results (see Fig.~\ref{contrast_1}).
Conversely, in order to obtain a plausible stitching result, we desire to define a perception-consistent energy function, such that the most invisible seam possesses the minimum energy.

\begin{figure*}[!t]
  \centering
  \includegraphics[width=0.85\textwidth]{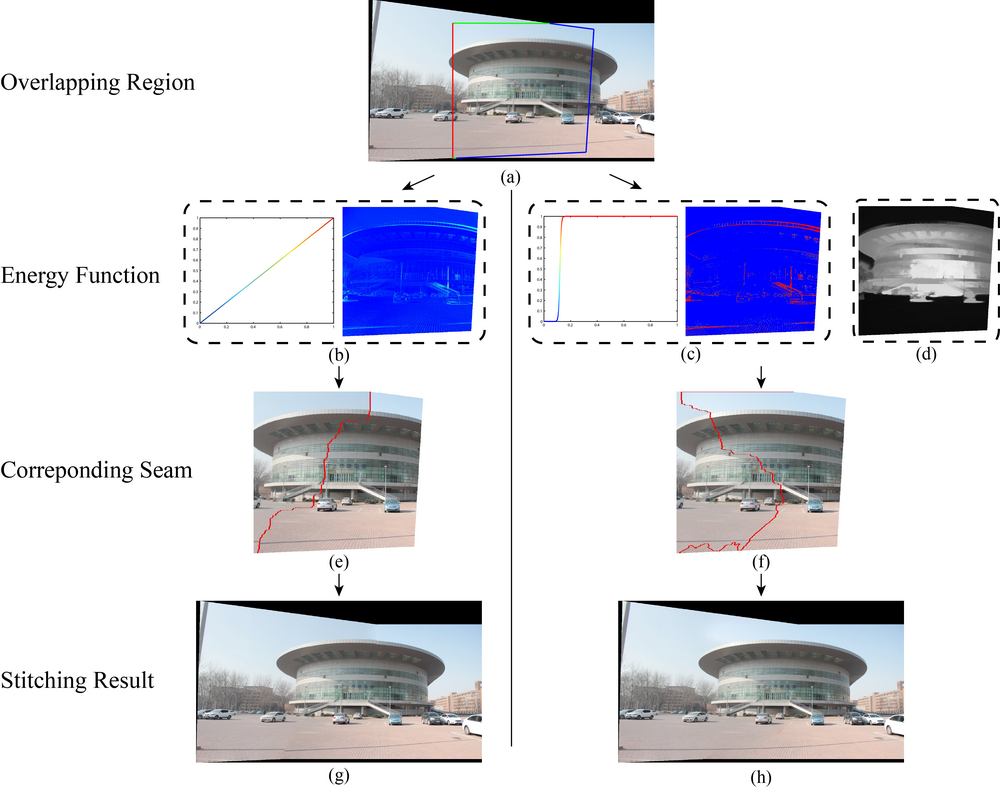}\\
  \caption{A process comparison between the normal seam-cutting framework and our proposed seam-cutting framework. (a) Overlapping region. (b) Euclidean-metric color difference. (c) Sigmoid-metric color difference. (d) Average pixel saliency. (e)(f) Corresponding seams. (g)(h) Corresponding stitching results.}
  \label{comp_sc}
\end{figure*}

Recently, many efforts have been devoted to seam-cutting by penalizing the photometric difference using various energy functions. A Euclidean-metric color difference is used in~\cite{kwatra2003graphcut} to define the smoothness term in their energy function, and a gradient difference is taken into account in~\cite{agarwala2004interactive}. Eden \emph{et al.}~\cite{Eden2006seamless} proposed an energy function that allows for large motions and exposure differences, but the camera setting is required. Jia and Tang~\cite{jia2008image} associated the smoothness term with gradient smoothness and gradient similarity, to reduce structure complexity along the seam. Zhang \emph{et al.}~\cite{zhang2016multi} combined alignment errors and a Gaussian-metric color difference in their energy function, to handle misaligned areas with similar colors. However, few of existing methods consider human perception in their energy functions, which sometimes causes that a seam with minimum energy is not most invisible in the overlapping region.

Seam-cutting has also been applied in image alignment. Gao \emph{et al.}~\cite{gao2013seam} proposed a seam-driven image stitching framework, which finds a best homography warp from some candidates with minimal seam costs instead of minimal alignment errors. Zhang and Liu~\cite{zhang2014parallax} combined homography and content-preserving warps to locally align images, where seam costs are used as a quality metric to predict how well a homography enables plausible stitching. Lin \emph{et al.}~\cite{lin2016seagull} proposed a seam-guided local alignment, which iteratively improves warping by adaptive feature weighting according to their distances to current seams.

In this paper, we propose a novel seam-cutting method via a perception-based energy function, which takes the nonlinearity and the nonuniformity of human perception into account. Our proposed method consists of three stages (see Fig.~\ref{comp_sc}). In the first stage, we calculate a sigmoid-metric color difference of the given overlapping region as the smoothness term, to characterize the perception of color discrimination. Then, we calculate an average pixel saliency of the given overlapping region as the saliency weight, to simulate that human eyes incline to pay
more attention to salient objects. Finally, we minimize the perception-based energy function by the graph-cut optimization, to obtain the seam and the corresponding composed result. Experiments show that our method outperforms the seam-cutting method of the normal energy function, and a user study demonstrates that our composed results are more consistent with human perception.

Major contributions of the paper are summarized as follows.
\begin{enumerate}
  \item We proposed a novel perception-based energy function in the seam-cutting framework.
  \item Our composition method can be easily implemented into other stitching pipelines.
\end{enumerate}

\section{Approach}

In this section, we first show more details of the normal seam-cutting framework, then a novel perception-based energy function is described, and finally we propose our seam-cutting framework.

\subsection{Normal Seam-cutting Framework}

Given a pair of aligned images denoted by $I_0$ and $I_1$, let $\mathcal{P}$ be their overlapping region and $\mathcal{L}=\{0,1\}$ be a label set, where ``$0$'' corresponds to $I_0$ and ``$1$'' corresponds to $I_1$, then a seam means assigning a label $l_p \in \mathcal{L}$ to each pixel $p \in \mathcal{P}$. The goal of seam-cutting is to find a labeling $l$ (i.e., a map from $\mathcal{P}$ to $\mathcal{L}$) that minimizes the energy function
\begin{equation}\label{basic_energy}
E(l) = \sum_{p\in \mathcal{P}}D_p(l_p)+\sum_{(p,q)\in \mathcal{N}}S_{p,q}(l_p,l_q),
\end{equation}
where $\mathcal{N}\subset\mathcal{P}\times\mathcal{P}$ is a neighborhood system of pixels. The \emph{data term} $D_p(l_p)$ represents the cost of assigning a label $l_p$ to a pixel $p\in\mathcal{P}$, and the \emph{smoothness term} $S_{p,q}(l_p,l_q)$ represents the cost of assigning a pair of labels $(l_p,l_q)$ to a pair of pixels $(p,q)\in\mathcal{N}$.

The data term is defined as
\begin{equation}\label{data}
  \left\{\begin{array}{ll}
           D_p(1)=0,~D_p(0)=\mu, & \mbox{ if } p\in\partial I_0\cap\partial\mathcal{P}, \\
           D_p(0)=0,~D_p(1)=\mu, & \mbox{ if } p\in\partial I_1\cap\partial\mathcal{P}, \\
           D_p(0)=D_p(1)=0, & \mbox{ otherwise,}
         \end{array}
  \right.
\end{equation}
where $\mu$ is a very large penalty to avoid mislabeling, $\partial I_k\cap\partial\mathcal{P}$ is the common border of $I_k$ ($k=0,1$) and $\mathcal{P}$ (marked in {\red red} and {\blue blue} respectively in Fig.~\ref{contrast_1}(a)). In fact, the data term $D_p(l_p)$ fixes the endpoints of the seam as the intersections of the two colored polylines.

The smoothness term is defined as
\begin{equation}\label{smoothness}
S_{p,q}(l_p,l_q) = \frac{1}{2}|l_p-l_q|(I_*(p)+I_*(q)),
\end{equation}
\begin{equation}\label{color_diff}
I_*(\cdot)=\|I_0(\cdot)-I_1(\cdot)\|_2,
\end{equation}
where $I_*(\cdot)$ denotes the Euclidean-metric color difference (see Fig.~\ref{comp_sc}(b)).

Finally, the normal energy function (\ref{basic_energy}) is minimized by graph-cut optimization~\cite{boykov2001fast} to obtain the seam (see Fig.~\ref{comp_sc}(e)) and the composed result (see Fig.~\ref{comp_sc}(g)). Obviously, the definition of the energy function plays the most important role in the seam-cutting framework.

\subsection{Perception-based Energy Function}

In experiments, the seam denoted by $l_*$, that minimizes the normal energy function (\ref{basic_energy}) is sometimes not most invisible in $\mathcal{P}$. In other words, there exists a seam denoted by $l_\dag$, that is more invisible but has a greater energy than $l_*$ (see Fig.~\ref{comp_sc} (e) and (f)). Therefore, we desire to define a perception-consistent energy function, such that the most invisible seam possesses the minimum energy.

\subsubsection{Sigmoid metric}

\begin{figure}[!t]
  \centering
  \subfloat[]{
  \includegraphics[width=0.115\textwidth]{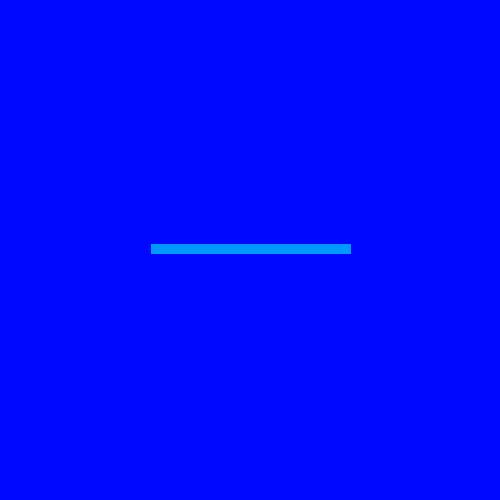}\label{toy_log_1}}
  \hfil
  \subfloat[]{
  \includegraphics[width=0.115\textwidth]{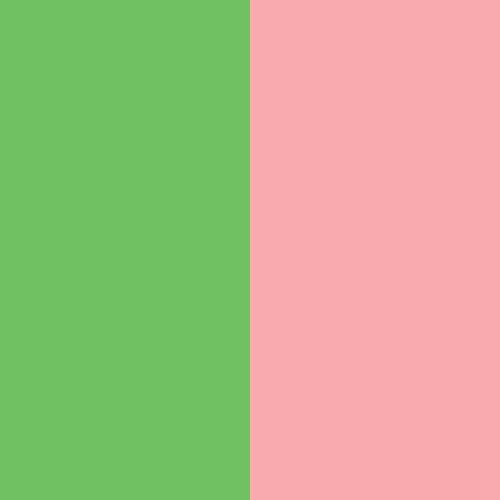}\label{toy_log_2}}
  \hfil
  \subfloat[]{
  \includegraphics[width=0.115\textwidth]{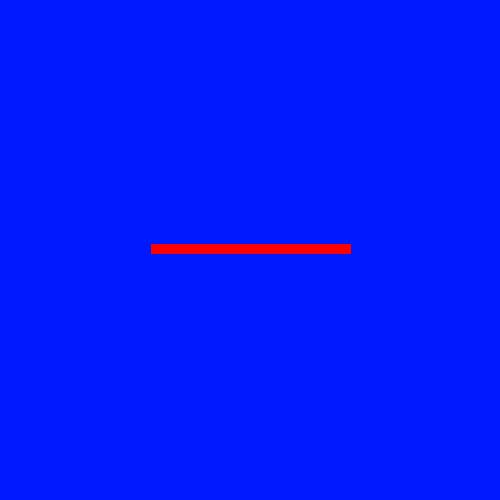}\label{toy_log_3}}
  \hfil
  \subfloat[]{
  \includegraphics[width=0.115\textwidth]{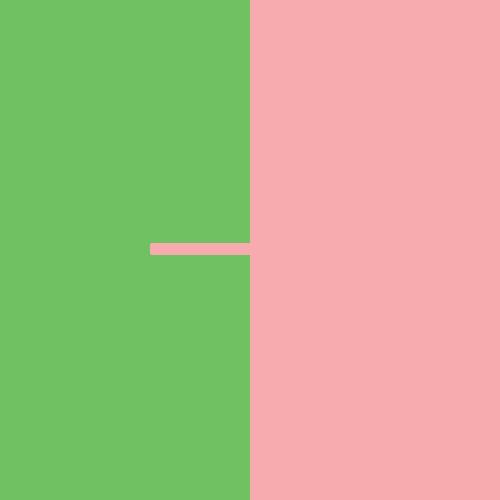}\label{toy_log_4}}
  \caption{Toy example. (a)(c) Visualizations of Euclidean-metric color difference and sigmoid-metric color difference. (b)(d) Corresponding seams.}
  \label{toy_log}
\end{figure}

Fig.~\ref{toy_log} shows a toy example where $l_*$ is not most invisible. In fact, the seam $l_*$ shown in (b) crosses the local misalignment area (marked in \textcolor[rgb]{0,0.6,1}{light blue} in (a)), because the Euclidean-metric color difference does not give it a large enough penalty. In contrast, the seam $l_\dag$ shown in (d) avoid the local misalignment area (marked in {\red red} in (c)), because the sigmoid-metric color difference successfully distinguish it from the alignment area.

In particular, the perception of colors is nonlinear as it has a color discrimination threshold, which means human eyes cannot differentiate some colors from others even if they are different. Let $\tau$ denote the threshold, the perception of color discrimination can be characterized as
\begin{itemize}
  \item if $I_*(\cdot)<\tau$, color difference is invisible,
  \item if $I_*(\cdot)\approx\tau$, sensitivity of discrimination rises rapidly,
  \item if $I_*(\cdot)>\tau$, color difference is visible.
\end{itemize}

We want to define a quality metric to measure the visibility of color difference, such that the cost of invisible terms approximates zero while the cost of visible terms approximates one. Fortunately, the sigmoid function
\begin{equation}
\mathrm{sigmoid}(x)=\frac{1}{1+e^{-4 \kappa(x-\tau)}},
\label{eq_logis}
\end{equation}
is a suitable quality metric for our purpose.

Next, we will show how to determine the parameters $\tau$ and $\kappa$. Briefly, given an overlapping region $\mathcal{P}$ of aligned images, the threshold $\tau$ plays the role of roughly dividing $\mathcal{P}$ into an alignment area and a misalignment area by its color difference, which is similar to determine a threshold to divide a binary image into a background region and a foreground region. Thus, we employ the well-known Ostu's algorithm~\cite{otsu1975threshold} to determine a suitable $\tau$ with the maximum between-class variance. On the other hand, $\kappa$ represents how rapidly the sensitivity of color discrimination rises around $\tau$. Normally, $\kappa=1/\epsilon$ will have a good practical performance, where $\epsilon$ is the width of bins of the histogram used in Ostu's algorithm.

Now, the smoothness term is modified as
\begin{equation}\label{proposed_smoothness}
\tilde{S}_{p,q}(l_p,l_q) = \frac{1}{2}|l_p-l_q|(I_\dag(p)+I_\dag(q)),
\end{equation}
\begin{equation}\label{new_color_diff}
  I_\dag(\cdot)=\mathrm{sigmoid}(I_*(\cdot)),
\end{equation}
where $I_\dag(\cdot)$ denotes the sigmoid-metric color difference. Fig.~\ref{comp_sc}(c) shows that $I_\dag(\cdot)$ forces the misalignment area more distinguishable from the alignment area than $I_*(\cdot)$, which effectively helps the seam avoid crossing the misalignment area.

\begin{figure}[!t]
  \centering
  \subfloat[]{
  \includegraphics[width=0.115\textwidth]{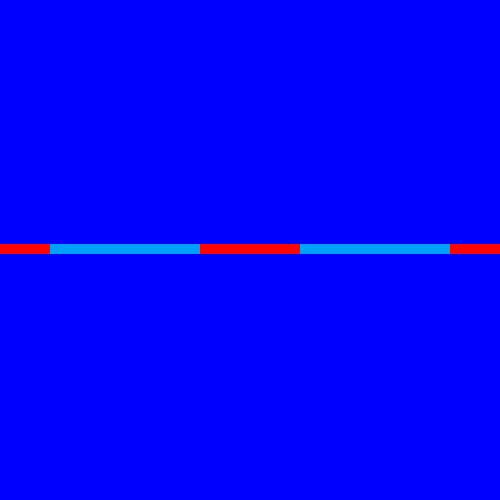}\label{toy_sal_1}}
  \hfil
  \subfloat[]{
  \includegraphics[width=0.115\textwidth]{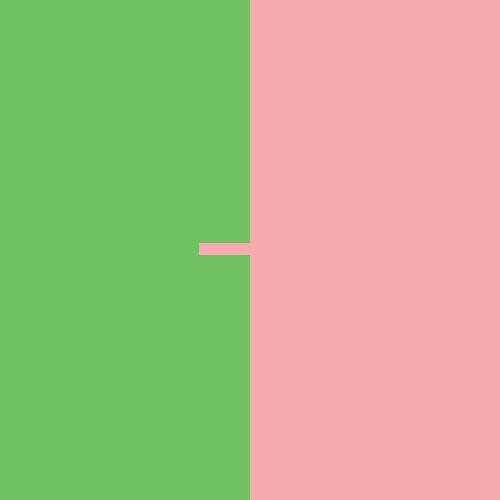}\label{toy_sal_2}}
  \hfil
  \subfloat[]{
  \includegraphics[width=0.115\textwidth]{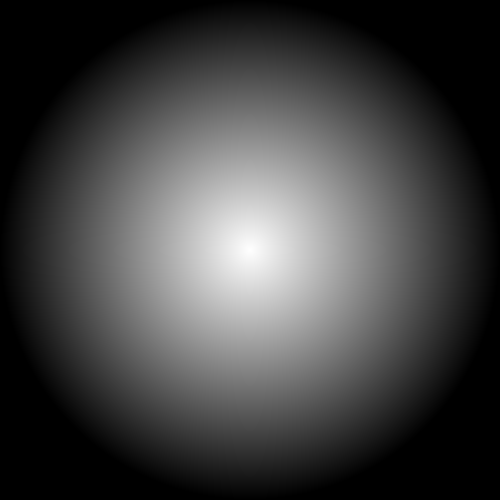}\label{toy_sal_3}}
  \hfil
  \subfloat[]{
  \includegraphics[width=0.115\textwidth]{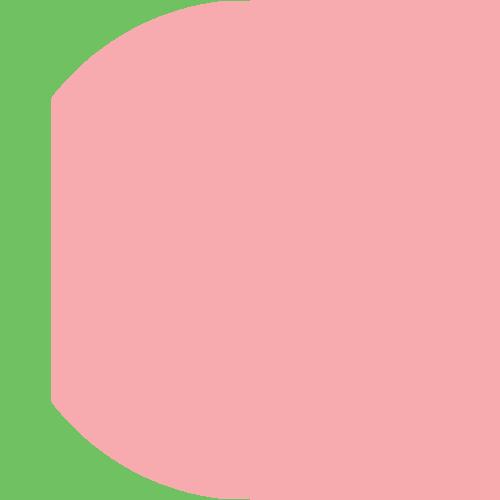}\label{toy_sal_4}}
  \caption{Toy example. (a) Visualization of sigmoid-metric color difference. (c) Visualization of average pixel saliency. (b)(d) Corresponding seams.}
  \label{toy_sal}
\end{figure}

\begin{figure*}[!t]
  \centering
  \includegraphics[width=0.88\textwidth]{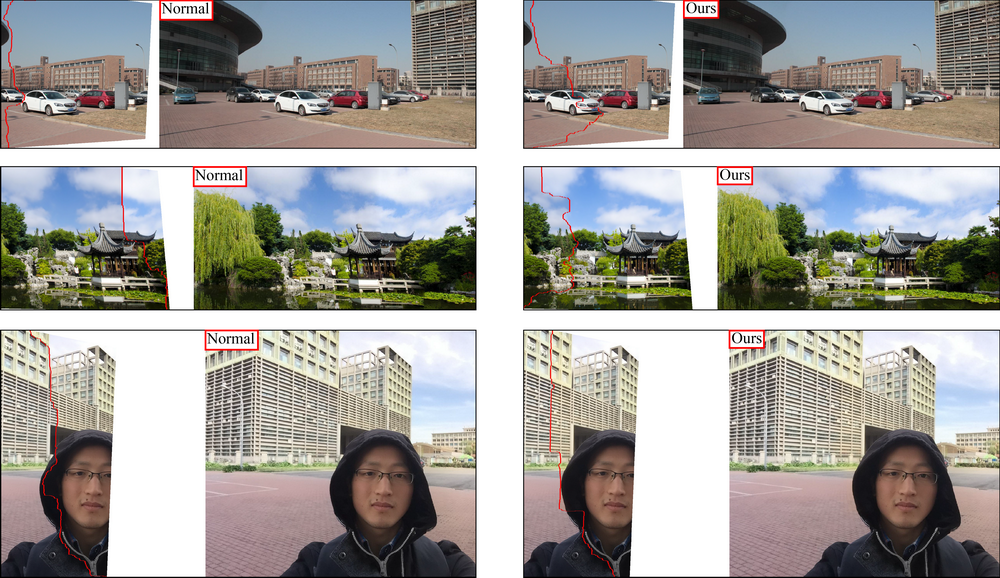}\\
  \caption{An experimental comparison between the normal seam-cutting framework and our perception-based seam-cutting framework. All stitching results are cropped into rectangles.}
  \label{exper_fig}
\end{figure*}

\subsubsection{Saliency weights}

Fig.~\ref{toy_sal} shows another toy example where $l_*$ is not most invisible. In fact, seams $l_*$ and $l_\dag$ shown in (b) and (d) respectively, both cross the local misalignment area. Though the energy of $l_\dag$ is greater, it is more invisible than $l_*$ in aspect of human perception, because the location where its artifact arises is less remarkable than $l_*$.

In particular, the perception of images is nonuniform, which means that human eyes incline to pay more attention to salient objects. Thus artifacts in salient regions are more remarkable than artifacts in non-salient regions. In order to benefit from these observations, we define a saliency weight
\begin{equation}
W_{p,q} = \left\{\begin{array}{ll}
0, & \mbox{if }p\,|\,q\in\partial_{\sharp} \mathcal{P},\\
1+\frac{\omega(p)+\omega(q)}{2},  & \mbox{otherwise,}
\end{array}\right.
\label{eq_sal}
\end{equation}
where $\omega(\cdot)$ denotes the average pixel saliency of $\mathcal{P}$ (see Fig.~\ref{comp_sc}(d)). We normalize $W_{p,q}$ in the range of $[1,2]$ to avoid over-penalizing saliency weights. As stitching results are usually cropped into rectangles in consumer-level photography, we assign $W_{p,q}=0$ if either $p$ or $q$ is located in the common border $\partial_{\sharp} \mathcal{P}$ of the canvas and $\mathcal{P}$ (marked in {\green green} in Fig.~\ref{comp_sc}(a)).

Finally, the perception-based energy function is defined as
\begin{equation}\label{proposed_energy}
\tilde{E}(l) = \sum_{p\in \mathcal{P}}D_p(l_p)+\sum_{(p,q)\in \mathcal{N}}W_{p,q}\cdot \tilde{S}_{p,q}(l_p,l_q),
\end{equation}
where $W_{p,q}$ rises the penalty of $\tilde{S}_{p,q}(l_p,l_q)$ according to $\omega(\cdot)$.
Fig.~\ref{comp_sc}(f) shows that the endpoints of the seam have more freedom on $\partial_{\sharp} \mathcal{P}$ than the seam shown in Fig.~\ref{comp_sc}(e).

\subsection{Proposed Seam-cutting Framework}

Our seam-cutting framework is summarized in Algorithm~\ref{algor_1}.

\begin{algorithm}
\caption{Perception-based seam-cutting framework.}
\label{algor_1}
\textbf{Input:} An overlapping region $\mathcal{P}$ of aligned images $I_0$ and $I_1$.\\
\textbf{Output:} A stitching result. \\
\vspace{-5pt}
\begin{enumerate}\setlength{\topsep}{0pt}
  \item Calculate $I_*(\mathcal{P})$ in Eq. (\ref{color_diff});
  \item Calculate $\tau$ in Eq. (\ref{eq_logis}) via Ostu's algorithm~\cite{otsu1975threshold};
  \item Calculate $I_\dag(\mathcal{P})$ in Eq. (\ref{new_color_diff}) and $\tilde{S}_{p,q}$ in Eq. (\ref{proposed_smoothness});
  \item Calculate $\omega(\mathcal{P})$ via salient object detection~\cite{zhang2015minimum} and $W_{p,q}$ in Eq. (\ref{eq_sal});
  \item Calculate $D_p(\mathcal{P})$ in Eq. (\ref{data});
  \item Minimize $\tilde{E}(l)$ in Eq. (\ref{proposed_energy}) via graph-cut optimization~\cite{boykov2001fast}, and blend $I_0$ and $I_1$ via gradient domain fusion~\cite{Perez:2003}.
\end{enumerate}
\end{algorithm}

\section{Experiments}

In our experiments, first, we use SIFT~\cite{lowe2004distinctive} to extract/match features, use RANSAC~\cite{fischler1981random} to determine a global homography and align input images. Then, for the overlapping region, we use Ostu's algorithm~\cite{otsu1975threshold} to estimate a threshold $\tau$ ($\epsilon=0.06$), and use salient object detection~\cite{zhang2015minimum} to calculate pixel saliency weights. Finally, we use graph-cut optimization~\cite{boykov2001fast} to obtain a seam, and blend aligned images via gradient domain fusion~\cite{Perez:2003} to create a mosaic.

Fig.~\ref{exper_fig} shows some experimental comparisons between two seam-cutting frameworks. Input images in the second group come from the dataset in \cite{zhang2014parallax}. Due to unconstrained shooting environment, there exist large parallax in these examples, such that a global homography can hardly align them. In such cases, the normal seam-cutting framework fails to produce artifact-free results, while our perception-based seam-cutting framework successfully creates plausible mosaics. More results and original input images are available in the supplementary material.

\begin{figure}
  \centering
  \includegraphics[width=0.35\textwidth]{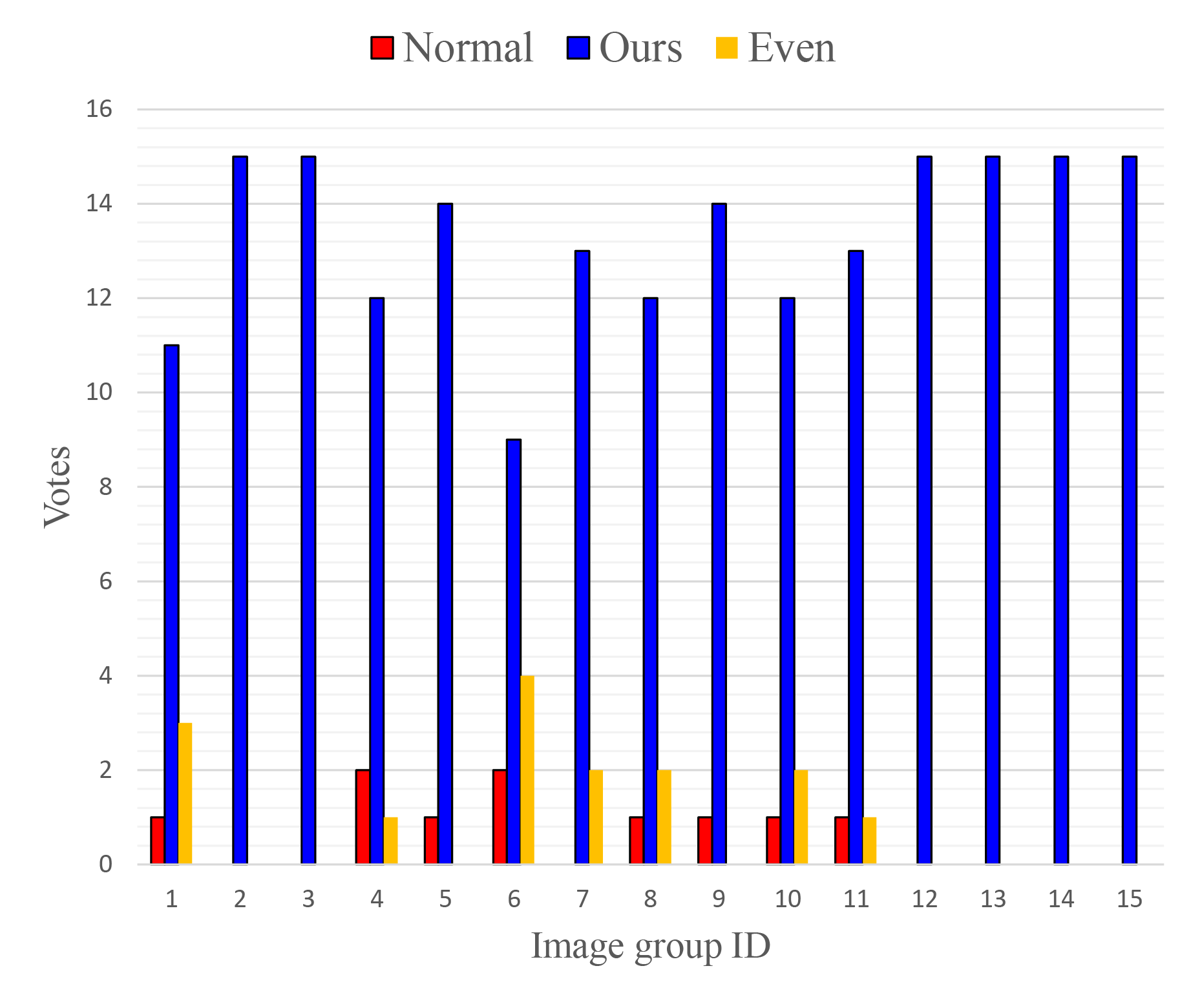}\\
  \caption{User study. {\red Red} represents the normal seam-cutting framework wins. {\blue Blue} represents our perception-based seam-cutting framework wins. {\textcolor[rgb]{1,0.76,0}{Yellow}} represents an even.}\label{user_study}
\end{figure}

In order to investigate whether our proposed method is more consistent with human perception, we conduct a user study for comparing two seam-cutting frameworks. We invite 15 participants to rank 15 unannotated groups of stitching results (make a choice from 3 options: 1. A is better than B, 2. B is better than A, 3. A and B are even). Fig.~\ref{user_study} shows the user study result, which demonstrates that our stitching results win most users' favor.

\section{Conclusion}

In this paper, we propose a novel perception-based energy function in the seam-cutting framework, to handle image stitching challenges in consumer-level photography. Experiments show that our method outperforms the seam-cutting method of the normal energy function, and a user study demonstrates that our results are more consistent with human perception. In the future, we plan to generalize our method in the seam-driven framework to deal with image alignment.

\ifCLASSOPTIONcaptionsoff
  \newpage
\fi

\newpage
\vfill
\balance

\end{document}